\documentclass[conference]{IEEEtran}
\IEEEoverridecommandlockouts
\usepackage{cite}
\usepackage{amsmath,amssymb,amsfonts}
\usepackage{algorithmic}
\usepackage{graphicx}
\usepackage{textcomp}
\usepackage[dvipsnames]{xcolor}

\usepackage[none]{hyphenat}
\usepackage{graphics} % for pdf, bitmapped graphics files
\graphicspath{{figures/}}
\usepackage[caption=false, font=normalsize, labelfont=sf, textfont=sf]{subfig}
\usepackage{physics}
\usepackage{textcomp}
\usepackage{mathtools}
\usepackage{blindtext}
\usepackage{tabularx}
\usepackage{multirow}
\usepackage{tabularx,booktabs}
\usepackage{lipsum}
\usepackage{cases}
\usepackage[short]{optidef}
\usepackage{url}
\usepackage[normalem]{ulem}
\usepackage[hidelinks]{hyperref}
\usepackage[linesnumbered,ruled,vlined]{algorithm2e}
\usepackage{soul}
\usepackage{tikz}

\newcommand{\cmmnt}[1]{}

\newcommand{\ie}{\textit{i.e.,}}

\newcommand{\etal}{\textit{et al.}}

\newcommand\copyrighttext{%
  \footnotesize \textcopyright 2025 IEEE. Personal use of this material is permitted. Permission from IEEE must be obtained for all other uses, in any current or future media, including reprinting/republishing this material for advertising or promotional purposes, creating new collective works, for resale or redistribution to servers or lists, or reuse of any copyrighted component of this work in other works.}
\newcommand\copyrightnotice{%
\begin{tikzpicture}[remember picture,overlay]
\node[anchor=north,yshift=-20pt] at (current page.north) {\fbox{\parbox{\dimexpr\textwidth-\fboxsep-\fboxrule\relax}{\copyrighttext}}};
\end{tikzpicture}%
}

\def\BibTeX{{\rm B\kern-.05em{\sc i\kern-.025em b}\kern-.08em
    T\kern-.1667em\lower.7ex\hbox{E}\kern-.125emX}}
\begin{document}

\copyrightnotice

\title{\LARGE \bf Biasing the Driving Style of an Artificial Race Driver\\for Online Time-Optimal Maneuver Planning*}

\author{Sebastiano Taddei$^{1,2}$, Mattia Piccinini$^{3}$, and Francesco Biral$^{1}$%
\thanks{*This work was partly supported by the European Union - Next Generation Eu - under the National Recovery and Resilience Plan (NRRP), Mission 4 Component 1 Investment 3.4 - Decree No. 351 of Italian Ministry of University and Research - Concession Decree No. 2152 of the Italian Ministry of University and Research, Project code D93C22000500001, within the Italian National Program PhD Programme in Autonomous Systems (DAuSy).}
\thanks{$^{1}$Sebastiano Taddei and Francesco Biral are with the Department of Industrial Engineering, University of Trento, 38123 Trento, Italy {\tt\small name.surname@unitn.it}}
\thanks{$^{2}$Sebastiano Taddei is also with the Department of Electrical and Information Engineering, Politecnico di Bari, 70125 Bari, Italy {\tt\small s.taddei@phd.poliba.it}}
\thanks{$^{3}$Mattia Piccinini is with the Professorship of Autonomous Vehicle Systems, Technical University of Munich, 85748 Garching, Germany {\tt\small mattia.piccinini@tum.de}}
}
% \title{\LARGE \bf Biasing the Driving Style of an Artificial Race Driver\\for Online Time-Optimal Maneuver Planning
% \thanks{This work was partly supported by the European Union Next-GenerationEU (Piano Nazionale di Ripresa e Resilienza (PNRR) - Missione 4 Componente 1, Investimenti 3.4 e 4.1 - Decreto del Ministero dell'Università e della Ricerca n.351 del 09/04/2022) within the Italian National Ph.D. Program in Autonomous Systems (DAuSy).}
% }

% \author{\IEEEauthorblockN{1\textsuperscript{st} Sebastiano Taddei}
% \IEEEauthorblockA{\textit{Department of Industrial Engineering} \\
% \textit{University of Trento}\\
% 38123 Trento, Italy \\
% sebastiano.taddei@unitn.it}
% \IEEEauthorblockB{\textit{Department of Electrical and Information Engineering} \\
% \textit{Politecnico di Bari}\\
% 70125 Bari, Italy \\
% s.taddei@phd.poliba.it}
% \and
% \IEEEauthorblockN{2\textsuperscript{nd} Mattia Piccinini}
% \IEEEauthorblockA{\textit{Department of Mobility Systems Engineering} \\
% \textit{TUM}\\
% Munich, Germany \\
% mattia.piccinini@tum.de}
% \and
% \IEEEauthorblockN{3\textsuperscript{rd} Francesco Biral}
% \IEEEauthorblockA{\textit{Department of Industrial Engineering} \\
% \textit{University of Trento}\\
% 38123 Trento, Italy \\
% francesco.biral@unitn.it}
% }

\maketitle

\begin{abstract}
In this work, we present a novel approach to bias the driving style of an artificial race driver (ARD) for online time-optimal trajectory planning. Our method leverages a nonlinear model predictive control (MPC) framework that combines time minimization with exit speed maximization at the end of the planning horizon. We introduce a new MPC terminal cost formulation based on the trajectory planned in the previous MPC step, enabling ARD to adapt its driving style from early to late apex maneuvers in real-time. Our approach is computationally efficient, allowing for low replan times and long planning horizons. We validate our method through simulations, comparing the results against offline minimum-lap-time (MLT) optimal control and online minimum-time MPC solutions. The results demonstrate that our new terminal cost enables ARD to bias its driving style, and achieve online lap times close to the MLT solution and faster than the minimum-time MPC solution. Our approach paves the way for a better understanding of the reasons behind human drivers' choice of early or late apex maneuvers.
\end{abstract}
\section{Introduction}
\label{sec:introduction}
Professional race car drivers are able to set minimum-lap-times by exploiting their driving skills and experience. As discussed in \cite{Kegelman2017}, human drivers may have different driving styles, yet achieving very similar lap times. This indicates the existence of many local minima in the space of minimum-lap-time maneuvers. Recently, \textit{artificial} drivers were developed, to compete in autonomous racing series \cite{Betz2022,mattia_piccinini_how_2024,Ogretmen2024,Piccinini2024_PathPolyNN} or become digital twins of professional drivers \cite{Ju2023}. However, open research questions remain on how to change the driving style of an artificial driver, while performing online (\ie{} real-time) trajectory planning.

In this paper, we aim to analyze and bias the driving style of an artificial race driver (ARD) for online time-optimal trajectory planning. We implement a nonlinear model predictive control (MPC) planner, which combines the time minimization and the exit speed maximization at the end of the planning horizon. Through a new terminal cost formulation, our ARD can be biased towards early- or late-apex maneuvers, and achieves online lap times very close to a minimum-lap-time offline solution.

\subsection{Motivation}
\label{sec:motivation}
Our motivation for biasing the driving style of artificial drivers is multifaceted. Human race drivers do not act like time-optimal MPC planners \cite{Anderson2016}, as they adapt their maneuvers online based on factors such as the geometry of the track that follows a corner, local execution errors, changing track and weather conditions, and whether they are in the lead or following an opponent.

Similarly, when performing MPC maneuver planning and execution close to the vehicle limits, adapting the MPC cost function terms in different corners can result in better overall performance \cite{anderson_modelling_2018}. Also, in autonomous racing, changing the driving style based on the current race scenario can help overtake other cars or adapt to varying track conditions. Finally, when
using an artificial race coach to train human drivers to improve their trajectories \cite{PagotPhDThesis}, the driving style of the artificial coach could be adapted to that of the human driver, and the coach could show the impact of different driving styles in different corners.

%These factors drove our work to bias the driving style of an artificial driver from early to late apex maneuvers.
%
\subsection{Apex Definition}
\label{sec:apex_definition}
Throughout this paper, we define the \emph{apex} as the point of a vehicle trajectory where the driver reaches the lowest speed (\ie{} stops decelerating and starts accelerating), and the \emph{clipping point} as the point of the trajectory that is closest to the inside of a corner.
\subsection{Related Work}
\label{sec:related_work}
The research of Anderson \etal{} \cite{Anderson2016,anderson_modelling_2018,anderson_cascaded_2020} recently showed that the trajectories of human race drivers can be modeled using receding horizon MPC, where the cost function combined time minimization and exit speed maximization. They used a genetic algorithm to locally tune the weights of the MPC cost function, and showed that the maximization of the exit speed improves the lap times and changes the planned maneuvers. Their approach is relevant, but has the following limitations. In \cite{Anderson2016}, they tracked a pre-computed minimum-time trajectory, and the MPC maneuver times were quite higher (1.7\%) than the optimal reference. In \cite{anderson_modelling_2018,anderson_cascaded_2020}, the MPC was computationally expensive (1 hour for a single corner) and thus not suitable for real-time planning, and they did not compare the results with minimum-lap-time optimal control.

Some authors, like \cite{Loeckel:2023aa,Werner2024,Doubek2021}, conducted surveys and interviews with professional race drivers, to understand their preferences and adaptation strategies in search of the minimum-lap-times and the optimal driving style.

Recently, a new stream of research has emerged around the use of imitation learning (IL) and reinforcement learning (RL) to model and imitate the driving styles of professional drivers. The authors of \cite{Ju2023,lockel2020probabilistic,Loeckel:2023aa,Lockel2022_identification} used IL and RL to reproduce and outperform the maneuvers of human drivers in simulation settings. Their results are relevant; however, they needed expert demonstrations to train their models.

Although the combination of time minimization and exit speed maximization has been studied for race driving style modeling \cite{Anderson2016,anderson_modelling_2018,anderson_cascaded_2020}, the existing literature is limited by at least by one of the following aspects:
\begin{itemize}
    \item High computational cost of the MPC problems, unsuited for online planning \cite{anderson_modelling_2018,anderson_cascaded_2020}.
    \item Tracking of pre-computed racelines and analysis of short track segments, rather than a full lap \cite{Anderson2016}.
    \item MPC maneuver times higher than a minimum-lap-time optimal control problem (MLT-OCP) \cite{Anderson2016}, or no comparison with it \cite{anderson_modelling_2018,anderson_cascaded_2020}.
    \item An MPC terminal cost was set to penalize the final states' deviations from an MLT-OCP, solved offline on a full lap \cite{piccinini_physics-driven_2023,Vazquez2020,Piccinini2022,subosits2019racetrack,mattia_piccinini_how_2024}. However, this makes the MPC dependent on the full-lap knowledge, requires to solve an MLT-OCP before the online MPC, and does not easily allow to change the driving style online.
\end{itemize}
\subsection{Contribution}
\label{sec:contribution}
Our contributions are as follows:
\begin{itemize}
    \item We develop an MPC-based artificial race driver that biases its driving style from early to late apex maneuvers, with low replan times for online operation and a long planning horizon.
    \item We devise a new MPC terminal cost based on the previous planned trajectory: our new terminal cost removes the need to solve offline a minimum-lap-time OCP and enables ARD to change its driving style online. This allows ARD to race in real-time without the need for a priori knowledge of the full track.
    \item We compare the results obtained with our approach against an offline minimum-lap-time optimal control problem and online pure minimum-time MPC solutions.
\end{itemize}
\section{Biasing the Driving Style}
\label{sec:biasing}
Most of the MPC-based racing trajectory planners in the literature aim to minimize the time to reach the end of the planning horizon \cite{piccinini_physics-driven_2023,Vazquez2020,mattia_piccinini_how_2024,Novi2020,Piccinini2020}. This typically results in a driving style that prefers early apexes, maximizing the use of combined lateral-longitudinal accelerations \cite{mattia_piccinini_how_2024} while minimizing the travelled distance whenever possible. However, such maneuvers often exploit regions of the combined lateral-longitudinal envelope where the vehicle exhibits major instabilities (\ie{} lower stability margins) that are unforgiving for driver errors. In contrast, recent analyses \cite{Werner2024,PagotPhDThesis,StefanLockel:2022aa} suggest that some professional drivers tend to prefer late apex maneuvers, which may be locally more stable and easier to execute. As we have seen in \cite{mattia_piccinini_how_2024,piccinini_physics-driven_2023}, artificial drivers also suffer from executions errors, which make them replan new time-optimal trajectories with locally wider paths, that can be executed without losing too much time. This suggests that, by biasing the driving style of an artificial driver towards late apexes, we may get more repeatable and human-like lap times.

In this work, we aim to bias the driving style of our MPC-based artificial race driver (ARD) by leveraging the maximization of the vehicle speed at the end of the horizon, and a new terminal cost formulation. We will show that our approach allows us to bias ARD's driving style from early to late apex, paving the way to a better understanding of the reasons behind the human driver's choice of early or late apex maneuvers.

Let us now describe our method, starting from the MPC problem formulation.

\subsection{Online Nonlinear Model Predictive Control}
\label{sec:enmpc}
The nonlinear MPC problem we use is adapted from our previous work \cite{piccinini_physics-driven_2023}, and is defined as follows:
\begin{equation}
    \label{eqn:enmpc}
    \min\limits_{\boldsymbol{u} \in \hspace{0.06cm} \mathcal{U}} \hspace{0.1cm} J \hspace{0.4cm} \text{s.t.} \hspace{0.1cm}
    \begin{cases}
        \dot{\boldsymbol{x}}(t) = \boldsymbol{f}(\boldsymbol{x}(t), \boldsymbol{u}(t)) \\
        \boldsymbol{b}(\boldsymbol{x}(0)) = 0 \\
        \boldsymbol{c}(\boldsymbol{x}(t), \boldsymbol{u}(t)) \geq 0
    \end{cases}
\end{equation}
where the cost function $J$ is:
\begin{align}
    \label{eqn:cost_function}
    J  &=  \overbrace{W_{n_{f}} \big(n(T) - n_{f}\big)^2+W_{\xi_{f}} \big(\xi(T) - \xi_{f}\big)^2}^{\text{terminal cost}} +\\\nonumber
    & -\overbrace{W_{v_x} v_x(T)}^{\text{exit speed}}
    + \overbrace{\int_{0}^{T} W_t \dd{t}}^{\text{maneuver time}}
\end{align}

The state vector $\boldsymbol{x}$ is defined as $\boldsymbol{x} = [n, \xi, v_x, v_y, \Omega, a_x]^T$. $n$ and $\xi$ are the lateral deviation and the relative yaw angle between the vehicle and the centerline, $v_x$ and $v_y$ are the longitudinal and lateral velocities, $\Omega$ is the yaw rate, and $a_x$ is the longitudinal acceleration. The control vector $\boldsymbol{u}$ is defined as $\boldsymbol{u} = [\Omega_0, a_{x_0}]^T$, where $\Omega_0$ is the desired yaw rate and $a_{x_0}$ is the desired longitudinal acceleration. The function $\boldsymbol{f}$ is a kineto-dynamical vehicle model, which is described in our previous work \cite{piccinini_physics-driven_2023}. Strict boundary conditions $\boldsymbol{b}$ are imposed on the initial states. The inequality constraints $\boldsymbol{c}$ impose the racetrack limits and the vehicle performance constraints, using a generalized polytopic formulation of the g-g-v diagram\footnote{The g-g-v diagram is a 3D map of $a_y$, $a_x$, and $v_x$, which encodes the maximum vehicle performance.} \cite{Piccinini_ggv_2024,mattia_piccinini_how_2024}.
Given that the final travel time $T$ is not known a priori, we reformulate the problem to use the path curvilinear abscissa $\zeta$ as the independent variable \cite{Lot2014}, instead of the time $t$. This allows us to solve the problem over a fixed horizon length $L$.

It is worth mentioning that the kineto-dynamical vehicle model $\boldsymbol{f}$ employed in our MPC is learned from experimental data, and does not rely on the prior knowledge of the vehicle dynamics or parameters.
This allows us to reproduce real vehicle\slash road characteristics at low computational costs, enabling real-time use. We describe the learning framework in \cite{piccinini_physics-driven_2023,mattia_piccinini_how_2024}.

The main novelties of our formulation are in the cost function \eqref{eqn:cost_function}, particularly in the exit speed maximization and the terminal cost. Let us describe these terms in detail.
\subsubsection{Maximizing the Exit Speed}
\label{sec:maximizing_vx}
The maximization of the longitudinal speed at the end of the horizon is obtained through the following term in the cost function \eqref{eqn:cost_function}:
\begin{equation}
    \label{eqn:vx_term}
    - W_{v_x} v_x(L)
\end{equation}
where $L$ is the length of the horizon. The weight $W_{v_x}$ is a positive scalar that determines the importance of the final longitudinal speed. By increasing the value of $W_{v_x}$, we can bias the driving style of our ARD from early to late apexes. To find a sensible range of values for $W_{v_x}$, we need to make sure that the value of \eqref{eqn:vx_term} is comparable to the maneuver time term in \eqref{eqn:cost_function}. A suitable value of $W_{v_x}$ can be estimated by assuming that the maximum speed $v_{x_{\text{max}}}$ is constant over a straight horizon. With this assumption, the maneuver time term in \eqref{eqn:cost_function} is at its minimum:
\begin{equation}
    \label{eqn:min_time}
    \int_{0}^{T} W_t \dd{t} = W_t \int_{0}^{L} \frac{\dd{\zeta}}{v_{x_{\text{max}}}} = W_t \frac{L}{v_{x_{\text{max}}}}
\end{equation}
and a suitable value of $W_{v_x}$ can be computed by equating the exit speed term in \eqref{eqn:vx_term} with \eqref{eqn:min_time}:
\begin{equation}
    \label{eqn:vx_weight}
    W_{v_x} v_{x_{\text{max}}} = W_t \frac{L}{v_{x_{\text{max}}}} \Longrightarrow W_{v_x} = W_t \frac{L}{v_{x_{\text{max}}}^2}
\end{equation}
Therefore, a sensible range of values for $W_{v_x}$ can be chosen around the guess given by (\ref{eqn:vx_weight}). We will discuss how the value in (\ref{eqn:vx_weight}) produces the best compromise, yielding the lowest MPC lap time.
\subsubsection{New Terminal Cost}
\label{sec:soft_final_conditions}
In the literature of minimum-time trajectory planning with MPC, a terminal cost is typically set to penalize the deviation of the final states from an MLT-OCP, solved on a full lap \cite{piccinini_physics-driven_2023,Vazquez2020,Piccinini2022,subosits2019racetrack,mattia_piccinini_how_2024}. This improves the MPC stability \cite{NMPC_book} and decreases the MPC computational times. However, solving an MLT is computationally expensive, and makes the MPC dependent on the full-lap knowledge, which may be a limitation. For example, if the maneuver execution error is too high, the terminal conditions from the MLT could be infeasible, which may yield MPC convergence issues.
Also, the MLT terminal cost would shadow the driving style bias we are trying to introduce with the maximization of the exit speed.

To overcome these limitations, we propose a new terminal cost formulation, based on the trajectory planned by the MPC in the previous step. Specifically, the terminal cost is imposed on the final states $n$ and $\xi$ (lateral displacement and relative yaw angle) in the cost function \eqref{eqn:cost_function} through the following term:
\begin{equation}
    W_{n_{f}} \big(n(L) - n_{f}\big)^2+W_{\xi_{f}} \big(\xi(L) - \xi_{f}\big)^2
\end{equation}
where $L$ is the horizon length. $n_{f}$ and $\xi_{f}$ are estimated by extrapolating over the new horizon the last planned trajectory, with the following logic.

\begin{figure}[ht]
    \centering
    \includegraphics[width=\columnwidth]{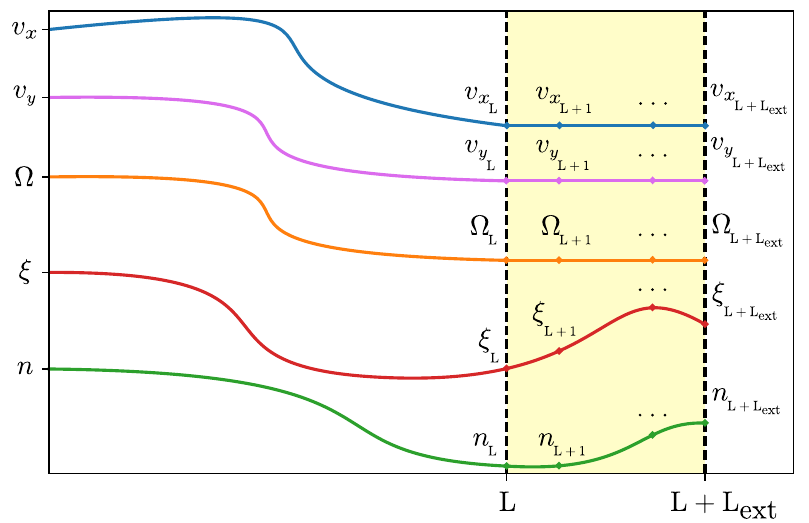}
    \caption{Extrapolation of the last planned trajectory to estimate the terminal conditions and state guess for the next MPC step. Between MPC steps, the previous solution needs to be shifted forward by $\textrm{L}_{\textrm{ext}}$ to estimate the terminal conditions and state guess for the next MPC step. $\textrm{L}$ is the end of the previous horizon, and $\textrm{L}_{\textrm{ext}}$ is the forward shift applied to the last planned trajectory. The extrapolation is performed from $\textrm{L}$ to $\textrm{L} + \textrm{L}_{\textrm{ext}}$. We can see that $v_x$, $v_y$, and $\Omega$ are assumed constant over the extrapolation horizon, while $n$ and $\xi$ are integrated forward using the dynamics \eqref{eqn:final_conditions} at $1\;\textrm{m}$ steps.}
    \label{fig:extrapolation}
\end{figure}

Let us consider the $n$ and $\xi$ dynamics expressed in curvilinear coordinates \cite{piccinini_physics-driven_2023,Lot2014}:
\begin{subnumcases}{\label{eqn:final_conditions}}
	\dv{n}{\zeta} = \frac{
        (v_x \sin{\xi} + v_y \cos{\xi}) \, (1 - n \, \kappa)
    }{
        v_x \cos{\xi} - v_y \sin{\xi}
    } \label{eqn:final_conditions_n} \\
    \dv{\xi}{\zeta} = \frac{
        \Omega \, (1 - n \, \kappa)
    }{
        v_x \cos{\xi} - v_y \sin{\xi}
    } - \kappa \label{eqn:final_conditions_xi}
\end{subnumcases}
where the dependency of all the variables on the curvilinear abscissa $\zeta$ is omitted, and $\kappa$ is the local road curvature. To estimate the terminal cost on $n$ and $\xi$, let us look at Fig. \ref{fig:extrapolation}. Between MPC steps, the previous solution needs to be shifted forward by $\textrm{L}_{\textrm{ext}}$ to estimate the terminal conditions and state guess for the next MPC step. $\textrm{L}$ is the end of the previous horizon, and $\textrm{L}_{\textrm{ext}}$ is the forward shift applied to the last planned trajectory. We assume that $v_x$, $v_y$, and $\Omega$ remain constant over the extrapolation horizon $L_{\textrm{ext}}$, and are equal to their values at the end of the previous horizon (\ie{} $v_x(L)$, $v_y(L)$, and $\Omega(L)$). To extrapolate $n$ and $\xi$, we start from their last values, $n(L)$ and $\xi(L)$, and integrate forward the dynamics \eqref{eqn:final_conditions} using an explicit Euler scheme with a step of $1\;\textrm{m}$ until we reach $n(L + L_{\textrm{ext}})$ and $\xi(L + L_{\textrm{ext}})$. This allows us to estimate their values at the end of the new horizon, which will be the terminal conditions for the next MPC step. In addition, their extrapolation over the new part of the horizon also proves useful as a state guess for the next MPC step, reducing the computational time of the MPC solver.

This terminal cost also has a physical meaning, which corresponds to estimating the knowledge of a human driver about the future track geometry, beyond the planning horizon.
Compared with the MLT-based terminal cost discussed at the beginning of this section, our approach allows the MPC to be independent of the full track knowledge and to change its driving style online. Thus, our MPC planner can start to operate without previously solving an MLT-OCP on the full track, which is an advantage over the existing literature \cite{piccinini_physics-driven_2023,Vazquez2020,Piccinini2022,subosits2019racetrack,mattia_piccinini_how_2024}.
Table \ref{tab:fc_comparison} shows that the lap time and the mean solve time of MPC with the traditional MLT-based and the proposed terminal cost are almost identical, differing by only $5\;\textrm{ms}$ and $3.953\;\textrm{ms}$, respectively.
This indicates that our approach is computationally efficient and can be used for real-time trajectory planning.
\begin{table}[ht]
    \centering
    \caption{Lap and mean solve times of MPC with different terminal costs. Unlike the MLT-based terminal cost, our cost does not require the full track knowledge, and enables the MPC to change its driving style online, without significantly affecting the performance.}
    \label{tab:fc_comparison}
    \begin{tabular}{ccc}
        \toprule
        & \multicolumn{1}{c}{MPC lap time} & \multicolumn{1}{c}{MPC mean solve time} \\
        \midrule
        MLT-based terminal cost & $113.551\;\textrm{s}$ & $24.970\;\textrm{ms}$ \\
        Proposed terminal cost & $113.556\;\textrm{s}$ & $28.923\;\textrm{ms}$ \\
        \bottomrule
    \end{tabular}
\end{table}
\subsection{Minimum-Lap-Time (MLT) on a Full Lap}
\label{sec:mlt}
As a baseline for our comparisons, we solve a minimum-lap-time (MLT) optimal control problem on a full circuit lap. The MLT problem has a similar formulation as the MPC problem \eqref{eqn:enmpc}, with the following differences. First, the cost function is only the maneuver time: we set $W_{v_x} = 0$ and $W_t = 2.0$. Then, we impose the initial and final states to be cyclic, \ie{} the vehicle states at the beginning and end of the lap must be the same. Lastly, we solve the MLT over one circuit lap. This yields the theoretical minimum-lap-time, given our problem formulation. We will use the MLT solution as a benchmark for our online MPC, whose cost function \eqref{eqn:cost_function} combines minimum-time and maximum exit speed.
\section{Results}
\label{sec:results}
\begin{figure*}[ht]
    \centering
    \includegraphics[width=\textwidth]{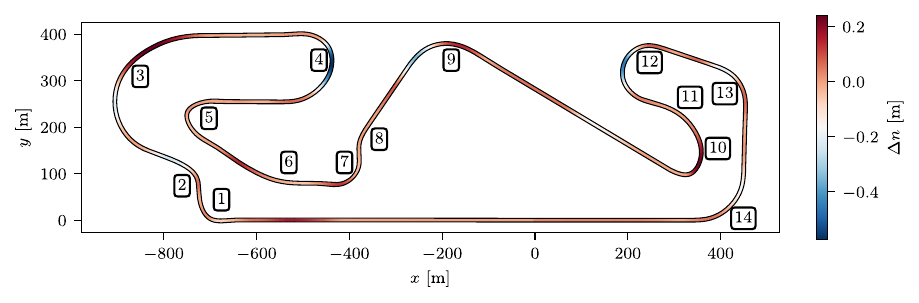}
    \caption{Difference between the lateral coordinates of $\textrm{MPC}_{0.00}$ and the best performing maximum-speed MPC, $\textrm{MPC}_{0.06}$ on the Catalunya Racetrack.}
    \label{fig:n_traj_diff}
\end{figure*}

The Catalunya racetrack, depicted in Fig. \ref{fig:n_traj_diff}, is the chosen track for our tests. The MPC planning horizon is set to $L = 300$ m, while the MPC trajectory is replanned every $50$ ms. In the MPC cost function \eqref{eqn:cost_function}, the minimum-time weight is set to $W_t = 2.0$, while the maximum-speed weight $W_{v_x}$ is varied from $0.0$ to $0.1$ in steps of $0.01$. Table \ref{tab:test_parameters} summarizes the MPC test parameters.

The entire software stack is implemented in \texttt{C++}, where we use the Pins software \cite{Biral2016} to formulate and solve the MPC and MLT problems. Pins is based on the indirect optimal control method, and was recently used for online minimum-time trajectory planning by \cite{piccinini_physics-driven_2023,Pagot2023Parking,Piccinini2020}. All tests are run on an M2 Max Apple Silicon chip.

In the next sections, we will show the drastic importance that a single weight ($W_{v_x}$) can have on the planned maneuvers, and the advantage of not using offline full-lap information in the MPC planning problem.

\begin{table}[ht]
    \centering
    \caption{MPC Parameters.}
    \label{tab:test_parameters}
    \begin{tabular}{ll}
        \toprule
        Parameter & Value \\
        \midrule
        % Track & Catalunya \\
        Planning horizon ($L$) & 300 m \\
        Replan time & 50 ms \\
        Minimum-time weight ($W_t$) & 2.0 \\
        Maximum-speed weight ($W_{v_x}$) & [0.00, 0.01, \dots, 0.09, 0.10] \\
        \bottomrule
    \end{tabular}
\end{table}

\subsection{Corner Analysis}
\label{sec:corner_analysis}
The MPC problem is solved online over the entire circuit, with different values of $W_{v_x}$, to analyze the differences in the planned maneuvers. In this section, the MPC solutions will be referred to as $\textrm{MPC}_{W_{v_x}}$, where $W_{v_x}$ will be substituted with the value of the weight used in the cost function. For example, $\textrm{MPC}_{0.00}$ corresponds to the pure minimum-time MPC solution ($W_{v_x} = 0.00$).

We will focus on the results obtained with $W_{v_x} = [0.00, 0.06, 0.10]$, as they are the most interesting configurations representing pure minimum-time planning, the best MPC lap time, and the maximum exit speed solutions, respectively. Section \ref{sec:trends} will analyze the sensitivity of the MPC solutions for more values of $W_{v_x}$. Let us now analyze the most interesting corners of the Catalunya racetrack.

\begin{figure*}[ht]
    \centering
    \includegraphics[width=\textwidth]{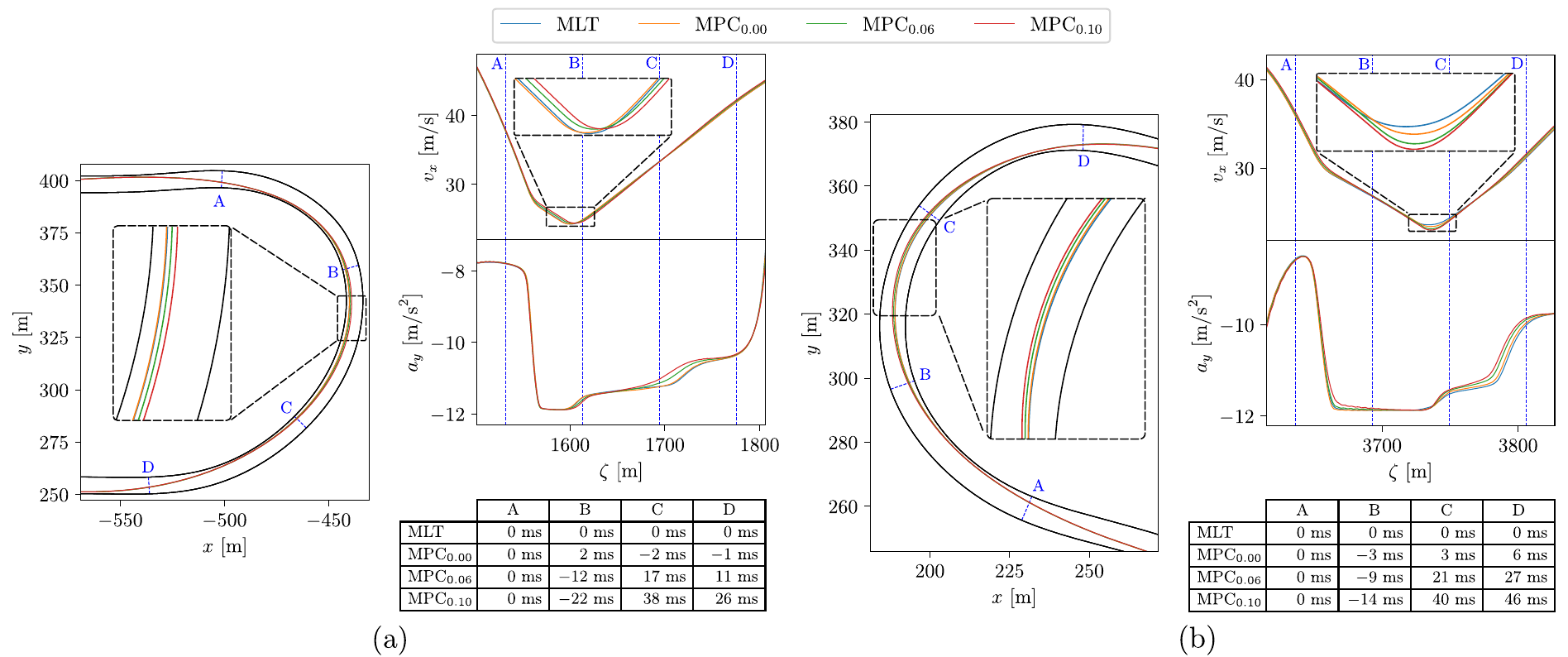}
    \caption{Corner n.4 (left) and Corner n.12 (right) of the Catalunya racetrack. Each figure, on the left, shows a zoom of the track segment used for the analysis, with four splits (A through D). On the right, it plots the longitudinal velocity and the lateral acceleration profiles over the four splits. In each figure, a table reports the travel time difference between the $\textrm{MPC}_{W_{v_x}}$ and the MLT solutions at the split points, where positive values indicate that the $\textrm{MPC}_{W_{v_x}}$ solution is locally slower than the MLT (the time counter starts at the beginning of the first split).}
    \label{fig:corner_4_12}
\end{figure*}

\begin{figure*}[ht]
    \centering
    \includegraphics[width=\textwidth]{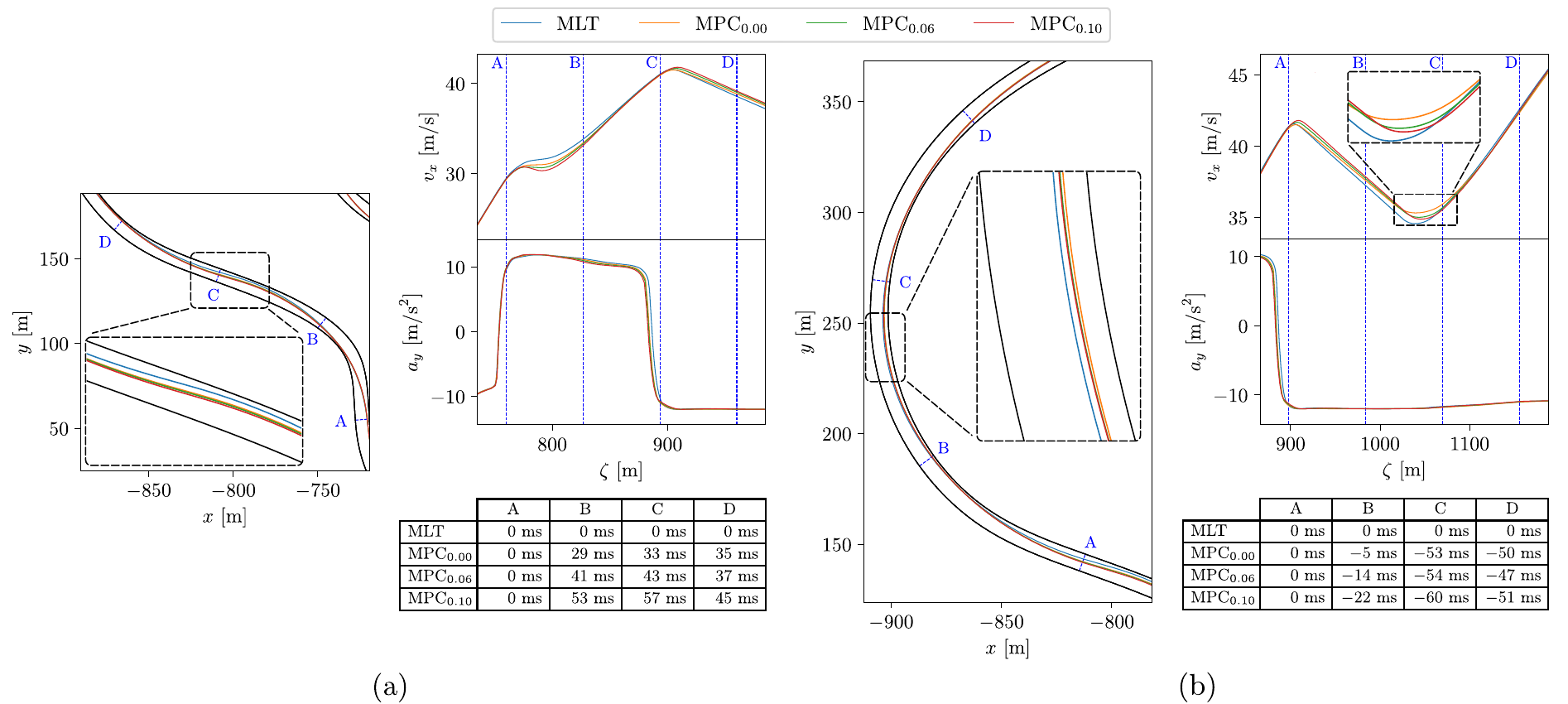}
    \caption{Straight segment (left) before Corner n.3 (right) of the Catalunya racetrack. For a detailed explanation, see the caption of Fig. \ref{fig:corner_4_12}.}
    \label{fig:corner_3}
\end{figure*}

Fig. \ref{fig:corner_4_12} and \ref{fig:corner_3} plot several corners of the Catalunya racetrack, to analyze how the planned trajectories differ due to the maximization of the exit speed. Each of the figures is structured as follows. On the left, it shows a zoom of the track segment used for the analysis, with four splits (A through D). On the right, it plots the longitudinal velocity and the lateral acceleration profiles over the four splits. In each figure, a table reports the travel time difference between the $\textrm{MPC}_{W_{v_x}}$ and the MLT solutions at the split points, where positive values indicate that the $\textrm{MPC}_{W_{v_x}}$ solution is locally slower than the MLT (the time counter starts at the beginning of the first split).
\paragraph*{Corner n.4}
Let us analyze the corner n.4 of the Catalunya racetrack, depicted in Fig. \ref{fig:corner_4_12}a. In the corner entry (sector A-B), the $\textrm{MPC}_{W_{v_x}}$ solutions carry more and more speed inside the turn and keep a progressively wider path as $W_{v_x}$ increases. Also, the $\textrm{MPC}_{W_{v_x}}$ keep a higher minimum speed, and their apexes (\ie{} the points of minimum speed) move towards the exit of the corner as $W_{v_x}$ increases (\textit{late apex}). This behavior is consistent with the maximization of the exit speed: for higher $W_{v_x}$, the $\textrm{MPC}_{W_{v_x}}$ keep a larger speed and a wider path in the corner exit. Also, the lateral acceleration $a_y$ is lower at the turn exit for higher $W_{v_x}$, which may suggest that these maneuvers are easier to execute. Conversely, the MLT stays closer to the inner track margin and keeps a lower speed throughout the corner, accelerating earlier (\textit{early apex}).

We will see how this behavior remains consistent over most corners.

The $\textrm{MPC}_{0.06}$ and $\textrm{MPC}_{0.10}$ solutions are quicker than the MLT in the A-B and C-D sectors, but they lose time along B-C, due to the locally wider paths.
\paragraph*{Corner n.12}
Corner n.12 (Fig. \ref{fig:corner_4_12}b) is another example of late apex biasing. Here, similar considerations hold as for corner n.4. The $\textrm{MPC}_{W_{v_x}}$ solutions with $W_{v_x} > 0$ keep a higher speed and a wider path up to their apexes. The minimum speed is lower for higher $W_{v_x}$, and the apexes move towards the turn exit. At the end of the corner (point D), the speed of the $\textrm{MPC}_{0.06}$ and $\textrm{MPC}_{0.10}$ solutions are higher than the MLT, due to the different cost functions. Once again, using $W_{v_x} > 0$ brings a reduction in $a_y$ at both the entry and exit of the corner, possibly easing the maneuver execution.
\paragraph*{Corner n.3}
Fig. \ref{fig:corner_3}b shows the corner n.3. Before looking at the corner itself, we need to analyze the straight before it, depicted in Fig. \ref{fig:corner_3}a. Along this short straight (segment B-C), the $\textrm{MPC}_{W_{v_x}}$ solutions choose a trajectory that gets wider and wider (towards the left track margin) as $W_{v_x}$ increases.
By doing so, in corner entry (point A in Fig. \ref{fig:corner_3}b) the $\textrm{MPC}_{W_{v_x}}$ brake later and from a higher speed, as $W_{v_x}$ increases. All maneuvers, including the MLT, have two clipping points, but the apex (\ie{} the point of minimum speed) moves towards the exit of the corner as $W_{v_x}$ increases (\textit{late apex}).
Conversely, the MLT trajectory is closer to the right track margin in the corner entry (Fig. \ref{fig:corner_3}a), but then brakes earlier and keeps a lower speed up to the acceleration point, which is reached before the $\textrm{MPC}_{W_{v_x}}$ (early apex).

Interestingly, the MLT is quicker than the $\textrm{MPC}_{W_{v_x}}$ in the corner entry (see the table in Fig. \ref{fig:corner_3}a), while the $\textrm{MPC}_{0.10}$ is the quickest through the corner (sector A-D in Fig. \ref{fig:corner_3}b). This shows how a late apex trajectory can be the time-optimal trajectory for a specific corner. Such a behavior is not possible with traditional minimum-time MPC, and shows the advantage of our approach, which allows a bias of the driving style.

\subsection{Summary and Lap Times}
The corner analysis shows that the trajectories planned by the $\textrm{MPC}_{W_{v_x}}$ are locally different from the MLT solution. The $\textrm{MPC}_{W_{v_x}}$ maneuvers follow a trend towards late apexes as $W_{v_x}$ increases. In corners, this yields wider trajectories with higher $v_x$ and lower $a_y$ in the entry and exit phases, possibly avoiding unstable regions of the performance envelope without sacrificing the travel time. Losing time over one corner may not be suboptimal: the MLT solution itself shows how sometimes trading off a few milliseconds on a corner may improve the overall lap time by allowing faster segments afterwards.

Table \ref{tab:lap_solve_time} compares the overall lap times of the different configurations of the $\textrm{MPC}_{W_{v_x}}$ problem and of the MLT. The MPC solution with $W_{v_x} = 0.06$ achieves the fastest lap time among the MPC configurations, confirming that our choice for the $W_{v_x}$ weight in \eqref{eqn:vx_weight} results is the best compromise between the minimum-time and maximum-speed terms. Notably, the $\textrm{MPC}_{0.06}$ solution is faster than the pure minimum-time $\textrm{MPC}_{0.00}$, which confirms the findings of \cite{anderson_modelling_2018}. The reason behind the $\textrm{MPC}_{0.06}$ being the fastest $\textrm{MPC}_{W_{v_x}}$ solution will be further discussed in the next section. Moreover, the lap time of the $\textrm{MPC}_{0.06}$ is only $7$ ms slower than the MLT, which is a negligible difference on a 4.5 km racetrack. This is an improvement of the results in \cite{Anderson2016}, and opens promising perspectives for local adaptations of the MPC weights, for example with a reinforcement learning method \cite{Zarrouki2021,Gottschalk2024}.
\begin{table}[ht]
    \centering
    \caption{Comparison of the lap and mean solve times of the MLT and $\textrm{MPC}_{W_{v_x}}$ problems.}
    \label{tab:lap_solve_time}
    \begin{tabular}{ccc}
        \toprule
        & \multicolumn{1}{c}{Lap time} & \multicolumn{1}{c}{MPC mean solve time} \\
        \midrule
        $\textrm{MLT}$        & $113.535\;\textrm{s}$ & - \\
        $\textrm{MPC}_{0.00}$ & $113.556\;\textrm{s}$ & $28.923\;\textrm{ms}$ \\
        $\textrm{MPC}_{0.06}$ & $113.542\;\textrm{s}$ & $27.682\;\textrm{ms}$ \\
        $\textrm{MPC}_{0.10}$ & $113.552\;\textrm{s}$ & $27.552\;\textrm{ms}$ \\
        \bottomrule
    \end{tabular}
\end{table}
\subsection{Sensitivity Analyses of the $\textrm{MPC}_{W_{v_x}}$ Solutions}
\label{sec:trends}
To understand the reason the $\textrm{MPC}_{0.06}$ is the fastest $\textrm{MPC}_{W_{v_x}}$ solution, let us perform a sensitivity analysis on the $\textrm{MPC}_{W_{v_x}}$ solutions.

\begin{figure}[ht]
    \centering
    \includegraphics[width=\columnwidth]{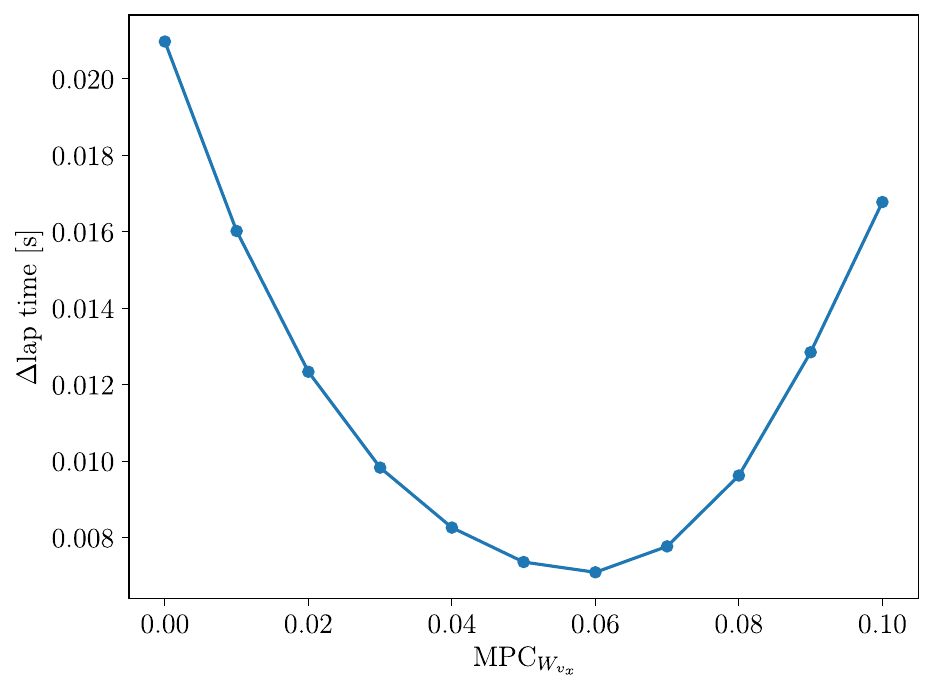}
    \caption{Difference between the lap times of the $\textrm{MPC}_{W_{v_x}}$ against the MLT solution. The $\textrm{MPC}_{0.06}$ is the fastest $\textrm{MPC}_{W_{v_x}}$ solution.}
    \label{fig:time_deltas}
\end{figure}

Fig. \ref{fig:time_deltas} plots the lap time differences between the $\textrm{MPC}_{W_{v_x}}$ and the MLT solution. We can see the time difference has a parabolic trend with a minimum at $W_{v_x} = 0.06$: this indicates that maximizing the exit speed initially gives a performance boost, but then it starts to worsen the lap time.

\begin{figure}[ht]
    \centering
    \includegraphics[width=\columnwidth]{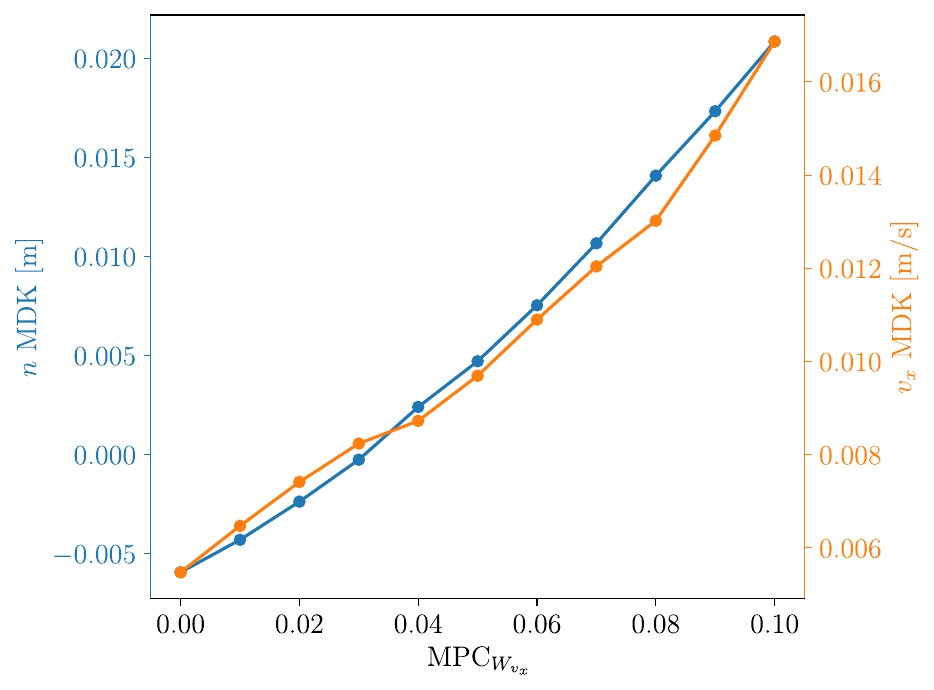}
    \caption{Mean deviation of the $\textrm{MPC}_{W_{v_x}}$'s lateral coordinates $n$ and longitudinal velocities $v_x$ with respect to the MLT solution: these mean deviations are weighted by the centerline curvature and normalized over the maximum curvature (MDK) (see eq. \eqref{eqn:mdk}). Positive values of $n$ deviation indicate wider paths in the corners, while positive values of $v_x$ deviation indicate higher speeds. As $W_{v_x}$ increases, both the lateral deviation and the longitudinal velocity deviation increase.}
    \label{fig:n_vx_md_kappa}
\end{figure}

Fig. \ref{fig:n_vx_md_kappa} shows the mean deviation of the $\textrm{MPC}_{W_{v_x}}$'s lateral coordinates $n$ and longitudinal velocities $v_x$ with respect to the MLT solution: these mean deviations are weighted by the centerline curvature and normalized over the maximum curvature (MDK). The MDK metric is computed as:
\begin{equation}
\label{eqn:mdk}
    \textrm{MDK} = \textrm{mean}\Big((x_{\textrm{MPC}_{W_{v_x}}} - x_{\textrm{MLT}}) \cdot \frac{\kappa} {\kappa_{\textrm{max}}}\Big)
\end{equation}
where $x$ is the state of interest and $\kappa$ the centerline curvature. Positive values of $n$ deviation indicate wider paths in the corners, while positive values of $v_x$ deviation indicate higher speeds. As $W_{v_x}$ increases, both the lateral deviation and the longitudinal velocity deviation increase. This trend highlights the impact of the $W_{v_x}$ weight in the cost function, and how it can bias the driving style of our ARD. We remark that our MPC is \textit{not} tracking the MLT solution, but rather re-planning online new trajectories with different driving styles.

\begin{figure}[ht]
    \centering
    \includegraphics[width=\columnwidth]{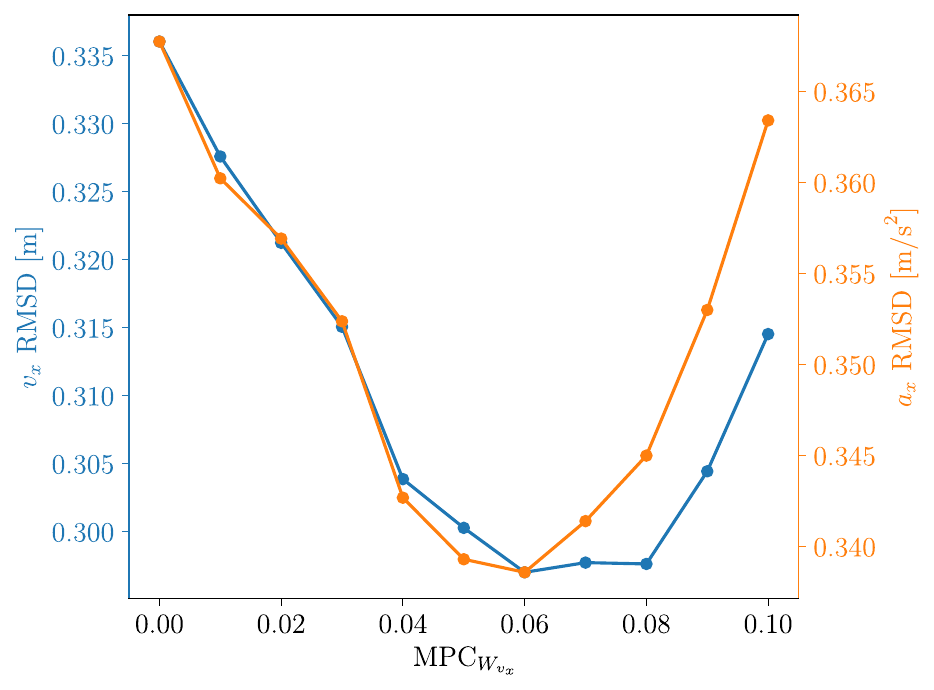}
    \caption{RMS deviation (RMSD) of $v_x$ and $a_x$ of the $\textrm{MPC}_{W_{v_x}}$ solutions from the MLT solution. Lower values of RMSD indicate a profile closer to the MLT solution. We can see that the $\textrm{MPC}_{0.06}$ has the lowest RMSD for both $v_x$ and $a_x$.}
    \label{fig:vx_ax_rmsd}
\end{figure}

Lastly, Fig. \ref{fig:vx_ax_rmsd} plots the root-mean-square deviation (RMSD) of $v_x$ and $a_x$ from the MLT solution. Lower values of RMSD indicate a profile closer to the MLT solution. The $\textrm{MPC}_{0.06}$ has the lowest RMSD for both $v_x$ and $a_x$, which suggests that the $\textrm{MPC}_{0.06}$ is the fastest solution because it is the closest to the MLT solution, while still preferring late apexes and maximizing speed in corners.


\subsection{Computational Performance}
\label{sec:performance}
Table \ref{tab:lap_solve_time} shows the mean computational time to solve a single $\textrm{MPC}_{W_{v_x}}$ problem. We can see that the mean time stays well below the replan-time of $50$ ms, ensuring real-time feasibility. It is worth noting that, compared to the related works \cite{anderson_modelling_2018,anderson_cascaded_2020}, with our approach we can easily solve the different configurations of the $\textrm{MPC}_{W_{v_x}}$ problem over the entire circuit in a handful of minutes. This allows us to quickly test different configurations and find the best compromise between the minimum-time and maximum-speed terms.
\section{Conclusions and Future Work}
\label{sec:conclusions}
In this paper, we presented a novel approach to bias the driving style of an artificial race driver (ARD) and to avoid the use of full-lap knowledge in the MPC. We showed how the maximization of the final longitudinal velocity, through a single parameter in the MPC cost function, can drastically change the driving styles. This final speed maximization leads to a driving style that prefers late apexes, which is more in line with the driving style of professional drivers. We compared the results obtained with our approach against both a minimum-lap-time optimal control solution and a minimum-time MPC solution. Our results suggest that maximizing the final longitudinal velocity not only drastically changes the planned trajectories, but also can lead to faster lap times. Additionally, we proved that we can avoid the use of full-lap knowledge in the MPC, without sacrificing lap performance nor computational time. Lastly, we showed that our approach is capable of real-time execution, allowing for quick testing of different configurations and real-world applications.

In future work, we plan to use reinforcement learning to dynamically adapt the weight $W_{v_x}$ corner by corner, to achieve faster lap times and\slash or mimic specific driving styles. Also, we will test our approach on different racetracks to evaluate its generalization capabilities.
\bibliographystyle{IEEEtran}
\bibliography{IEEEabrv,biblio}
\end{document}